\documentclass[letterpaper,10pt,conference]{ieeeconf}

\IEEEoverridecommandlockouts %
\usepackage{times}    %
\usepackage{amsmath}  %
\usepackage{amssymb}  %
\usepackage{bm}       %
\usepackage{booktabs}

\usepackage{array}
\usepackage{epsfig}
\usepackage{etoolbox}
\usepackage{graphics}
\usepackage{makecell}
\usepackage{microtype}
\usepackage{stfloats}

\usepackage{mathtools}  %
\DeclarePairedDelimiter\abs{\lvert}{\rvert}%

\usepackage[noadjust]{cite} %
\usepackage{subcaption}
\usepackage[bookmarks=true,hidelinks]{hyperref}
\usepackage{paralist}  %

\usepackage[binary-units=true]{siunitx}

\newcommand{\vq}{\bm{q}}

\newcommand{\vx}{\bm{x}}

\newcommand{\refx}{\vx^{(0)}}
\newcommand{\refu}{\vu^{(0)}}

\newcommand{\vu}{\bm{u}}
\newcommand{\vv}{\bm{v}}
\newcommand{\vf}{\bm{f}}
\newcommand{\vk}{\bm{k}}

\newcommand{\argmin}{\operatornamewithlimits{arg\ min}}

\newcommand{\delx}{\delta \vx}
\newcommand{\delu}{\delta \vu}

\usepackage[latin1]{inputenc}

\usepackage{graphicx}
\graphicspath{{figures/}}
\usepackage[export]{adjustbox} %

\usepackage[style=long,nolist]{glossaries}
\glsdisablehyper
\newacronym{sparsefddp}{SparseFDDP}{Sparsity-inducing feasibility-driven Dynamic Differential Programming}

\newacronym{c-space}{$\mathcal{C}$-space}{configuration space}
\newacronym{com}{CoM}{Centre-of-Mass}
\newacronym{dof}{DoF}{degrees of freedom}
\newacronym{eom}{EoM}{equation of motion}
\newacronym{fk}{FK}{forward kinematics}
\newacronym{giw}{GIW}{Gravito-Inertial Wrench}
\newacronym{ik}{IK}{inverse kinematics}

\newacronym{hmm}{HMM}{Hidden Markov Model}

\newacronym{lq}{LQ}{Linear-Quadratic}
\newacronym{lp}{LP}{linear program}
\newacronym{lping}{LP}{Linear Programming}
\newacronym{nlp}{NLP}{Non-linear Programming}
\newacronym{qp}{QP}{Quadratic Programming}
\newacronym{sdp}{SDP}{Semidefinite Programming}
\newacronym{sqp}{SQP}{Sequential Quadratic Programming}
\newacronym{miqcqp}{MIQCQP}{Mixed-Integer Quadratically Constrained Quadratic Programme}
\newacronym{sip}{SIP}{Semi-Infinite Programming}

\newacronym{lbfgs}{L-BFGS}{Limited-memory BFGS}
\newacronym{bfgs}{BFGS}{Broyden-Fletcher-Goldfarb-Shanno}
\newacronym{prm}{PRM}{Probabilistic Roadmap}
\newacronym{rrt}{RRT}{Rapidly-Exploring Random Tree}
\newacronym{hdrm}{HDRM}{Hierarchical Dynamic Roadmap}
\newacronym{drm}{DRM}{Dynamic Roadmap}
\newacronym{gjk}{GJK}{Gilbert--Johnson--Keerthi}
\newacronym{epa}{EPA}{Expanding Polytope Algorithm}
\newacronym{cmaes}{CMA-ES}{Covariance Matrix Adaptation Evolution Strategy}
\newacronym{pso}{PSO}{Particle Swarm Optimization}
\newacronym{chomp}{CHOMP}{Covariant Hamiltonian Optimization for Motion Planning}
\newacronym{stomp}{STOMP}{Stochastic Trajectory Optimization for Motion Planning}
\newacronym{aico}{AICO}{Approximate Inference COntrol}
\newacronym{riemo}{RieMo}{Riemannian Motion Optimization}
\newacronym{komo}{KOMO}{k-Order Motion Optimization}
\newacronym{pi2}{$\text{PI}^2$}{Policy Improvement with Path Integrals}
\newacronym{trajopt}{TrajOpt}{Trajectory Optimization for Motion Planning}
\newacronym{ddp}{DDP}{Differential Dynamic Programming}
\newacronym{sbp}{SBP}{Sampling-based Planning}
\newacronym{cg}{CG}{Conjugate Gradient}
\newacronym{knn}{KNN}{$k$-Nearest Neighbor}
\newacronym{pca}{PCA}{Principal Component Analysis}
\newacronym{mse}{MSE}{Mean Squared Error}
\newacronym{mae}{MAE}{Mean Absolute Error}
\newacronym{lwpr}{LWPR}{Locally Weighted Projection Regression}
\newacronym{gpr}{GPR}{Gaussian Process Regression}

\newacronym{admm}{ADMM}{Alternating Direction Method of Multipliers}
\newacronym{mpc}{MPC}{Model-Predictive Control}
\newacronym{pbd}{PbD}{Programming by Demonstration}
\newacronym{lfd}{LfD}{Learning from Demonstration}
\newacronym{ioc}{IOC}{Inverse Optimal Control}
\newacronym{irl}{IRL}{Inverse Reinforcement Learning}
\newacronym{oc}{OC}{Optimal Control}
\newacronym{rl}{RL}{Reinforcement Learning}

\newacronym{moe}{MoE}{Mixture-of-Experts}
\newacronym{poe}{PoE}{Product-of-Experts}
\newacronym{gmm}{GMM}{Gaussian Mixture Model}

\newacronym{scd}{SCD}{Smooth Collision Distance}
\newacronym{sop}{SoP}{Sum-of-Penetrations}
\newacronym{cd}{CD}{Collision Distance}
\newacronym{cc}{CC}{Collision Check}

\newacronym{sme}{SME}{Small and Medium Enterprise}
\newacronym{agv}{AGV}{Autonomous Ground Vehicle}
\newacronym{ros}{ROS}{Robot Operating System}
\newacronym{imu}{IMU}{Inertial Measurement Unit}
\newacronym{gps}{GPS}{Global Positioning System}
\newacronym{slam}{SLAM}{Simultaneous Localisation and Mapping}
\newacronym{ukf}{UKF}{Unscented Kalman Filter}
\newacronym{rsi}{RSI}{Repetitive Strain Injury}
\newacronym{oem}{OEM}{Original Equipment Manufacturer}
\newacronym{drc}{DRC}{{DARPA} Robotics Challenge}
\newacronym{eod}{EOD}{Explosive Ordnance Disposal}
\newacronym{idrm}{iDRM}{inverse Dynamic Reachability Map}
\newacronym{mit}{MIT}{Massachusetts Institute of Technology}
\newacronym{exotica}{EXOTica}{Extensible Optimization Toolset}

\usepackage[english]{babel}

\hypersetup{
    pdfauthor   = {Traiko Dinev and Wolfgang Merkt},%
    pdftitle    = {Sparsity-Inducing Optimal Control via Differential Dynamic Programming},%
    pdfsubject  = {Robotics},%
    pdfkeywords = {Robots, Trajectory Optimization, Differential Dynamic Programming, Sparsity, Dynamics}%
}

\title{\LARGE \bf
    Sparsity-Inducing Optimal Control via Differential Dynamic Programming
    }

\author{
    Traiko Dinev$^*$, Wolfgang Merkt$^*$, Vladimir Ivan, Ioannis Havoutis, Sethu Vijayakumar
    \thanks{$^*$ Equal contribution, alphabetical order.}%
    \thanks{Wolfgang Merkt and Ioannis Havoutis are with the Oxford Robotics Institute, University of Oxford, UK.}%
    \thanks{Traiko Dinev, Vladimir Ivan, and Sethu Vijayakumar are with the Edinburgh Centre for Robotics, The University of Edinburgh, UK.}%
    \thanks{This research was supported by (1) the European Commission under the Horizon 2020 project Memory of Motion (MEMMO, ID: 780684) and (2) the Engineering and Physical Sciences Research Council (EPSRC) UK RAI Hubs for Offshore Robotics for Certification of Assets (ORCA, EP/R026173/1) and Future AI and Robotics for Space (FAIR-SPACE, EP/R026092/1).}%
}

\begin{document}
\bstctlcite{IEEEexample:BSTcontrol}

\maketitle
\thispagestyle{empty}
\pagestyle{empty}

\begin{abstract}
    Optimal control is a popular approach to synthesize highly dynamic motion. Commonly, $L_2$ regularization is used on the control inputs in order to minimize energy used and to ensure smoothness of the control inputs. However, for some systems, such as satellites, the control needs to be applied in sparse bursts due to how the propulsion system operates.
    In this paper, we study approaches to induce sparsity in optimal control solutions---namely via smooth $L_1$ and Huber regularization penalties.
    We apply these loss terms to state-of-the-art \gls{ddp}-based solvers to create a family of sparsity-inducing optimal control methods.
    We analyze and compare the effect of the different losses on inducing sparsity, their numerical conditioning, their impact on convergence, and discuss hyperparameter settings.
    We demonstrate our method in simulation and hardware experiments on canonical dynamics systems, control of satellites, and the NASA Valkyrie humanoid robot.
    We provide an implementation of our method and all examples for reproducibility on GitHub.
\end{abstract}

\section{Introduction}
The propulsion systems of orbital satellites have a unique control limitation. In many cases, they use impulsive cold gas or bi-propellant thrusters incapable of or ineffective at low rates of firing. The resulting control then relies on fewer longer bursts of thrust to generate a constant amount of force over time while keeping the thrusters off between the bursts. 

The required control inputs are then a sequence of binary on/off commands that are activated sparsely throughout the motion~\cite{malyuta_fast_2020}. We call this type of control \emph{sparsity-inducing}, referring to the sparse use of control inputs throughout the trajectory. Such control will prefer zero control (off) followed by a high control action (on) to continuous corrective commands applied throughout the whole trajectory.

Selecting the required \glspl{dof} of a redundant system such as a humanoid robot is another application of \emph{sparsity-inducing} control. Instead of deactivating the control inputs, the planner deactivates unnecessary joints.
Consider a reaching task for the 38-\gls{dof} Valkyrie humanoid robot, where the goal is to extend the hand (end-effector) forward to point at a target. Solving this via motion planning involves finding suitable control inputs for the entire humanoid such that it balances itself and extends the hand. With traditional control-penalty methods, the planner readily discovers a motion where all the joints are simultaneously moving (cf. \autoref{fig:intro}). Such a motion is arguably unnecessary and in fact undesirable as it requires more complex control to coordinate the motion. This can result in trajectories that are more difficult to execute and track by the low-level controller and may require more energy.

To achieve sparsity in the controls as well as automatic joint selection, we propose to use a sparsity-inducing penalty term in the control cost. This serves to both switch off unnecessarily small control inputs in the case of high-\gls{dof} robots and to discover thruster-like behavior for satellites.

\begin{figure}
    \begin{minipage}{0.27\textwidth}
        \centering
        \includegraphics[width=\textwidth]{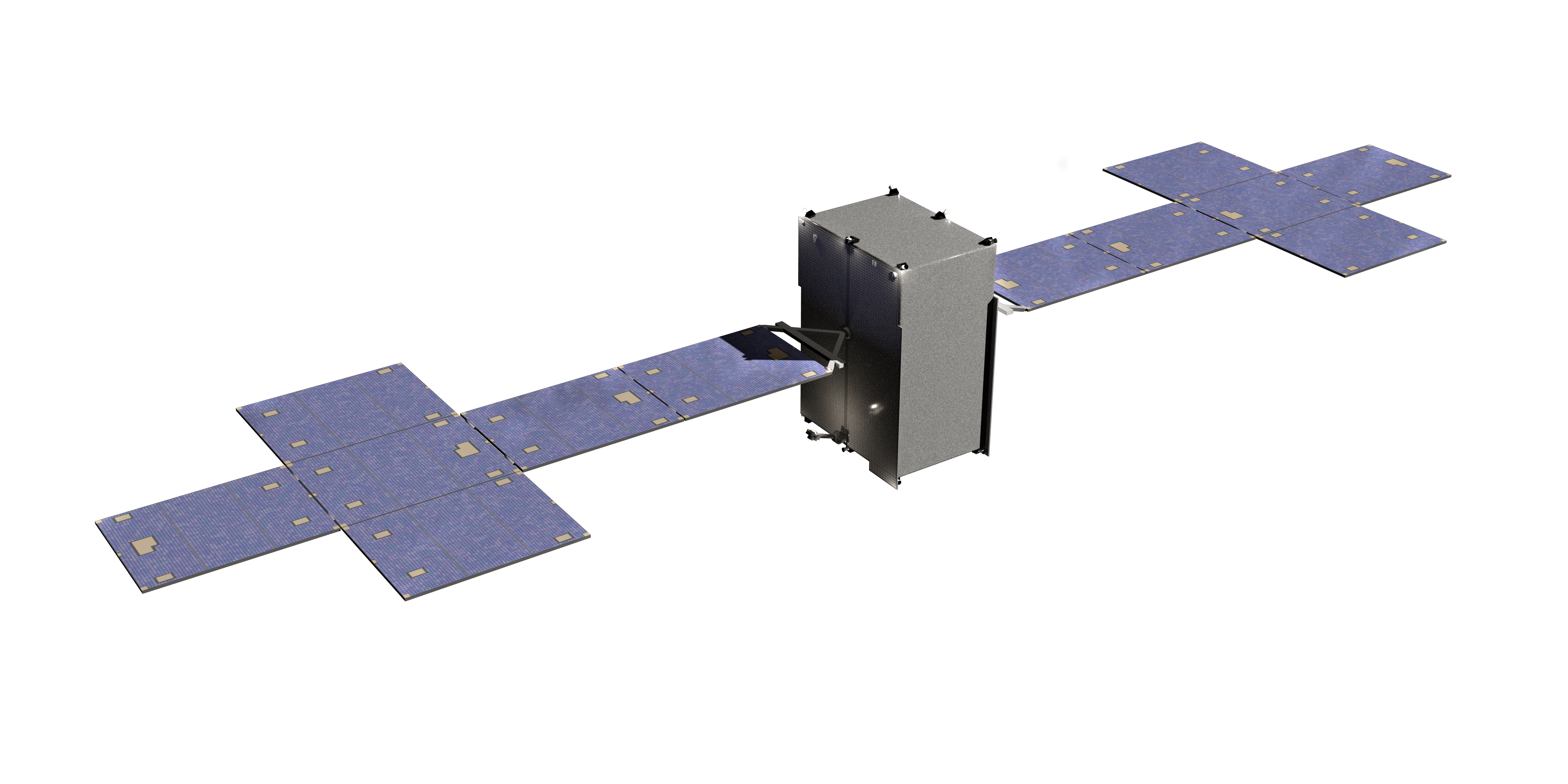}
    \end{minipage}
    \hfill
    \begin{minipage}{0.21\textwidth}
        \centering
        \includegraphics[width=\textwidth]{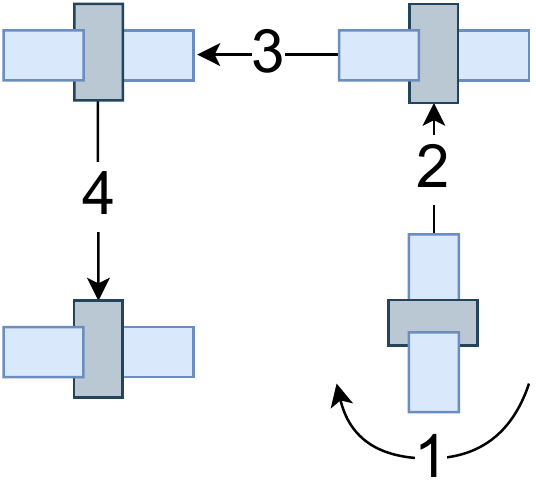}
    \end{minipage}
    \caption{SSL-1300 satellite model and satellite maneuver.}
    \label{fig:satellite-maneuver}
    \vspace{-1em}
\end{figure}

\subsection{Related Work}

Optimal control methods synthesize dynamically consistent motion satisfying a set of task and dynamics constraints while minimizing an optimality criterion. 
Shooting methods, which optimize over the control inputs, and in particular algorithms derived from \acrfull{ddp}~\cite{mayne1966ddp}, have recently received renewed interest~\cite{giftthaler_family_2017, neunert_trajectory_2017, mastalli_crocoddyl_2020}. In contrast to simultaneous transcription methods~\cite{kelly_trajectory_2017,ferrolho2020optimizing}, shooting methods have faster computation times by explicitly exploiting the temporal structure and implicitly enforcing the dynamic feasibility of the solution.

A common optimality criterion is energy optimization, which is traditionally a squared cost on the control inputs ($L_2$ norm regularization).
In practice, this leads to smooth control profiles and has been widely applied to canonical dynamic systems, computation of flight trajectories, as well as to synthesize highly dynamic maneuvers for legged robots \cite{neunert_trajectory_2017,mastalli_crocoddyl_2020}.
However, as no sparsity is introduced, on redundant systems this frequently leads to moving many joints even if not all joints are required to complete the task (cf. \autoref{fig:intro}).

Sparsity in control inputs for planning was studied in the context of satellite motion planning~\cite{hartley_terminal_2013}. There the authors used the $L_1$ norm, also known as \emph{Lasso} model. Similarly, the authors in \cite{cleach_fast_2019} applied an $L_1$ penalty using an \gls{admm} approach separating the problem into an optimal control update and a soft thresholding update. Whereas previous work considered an $L_1$ penalty applied to the force at the center of mass of the satellite, here we model the thruster behavior directly. Finally, the authors in \cite{malyuta_fast_2020} obtained thruster controls by optimizing the timing of thruster pulses, which are modeled as on-off controls. Here we use smooth $L_1$ costs to penalize the otherwise continuous thruster forces.
As a result, our approach is directly applicable in standard optimal control frameworks.

Recently, the concept of sparsity has gained additional attention in the trajectory optimization and control community for terrestrial/traditional robotics applications such as manipulation. \cite{hoffman2020study}, for instance, investigated the use of Mixed-Integer and Lasso regression to reduce joint motion on humanoid robots in a hierarchical inverse dynamics control scheme.
Nonetheless, enforcing sparsity for planning over longer horizons continues to be a challenge.

It is well known in the machine learning community that $L_1$ introduces a discontinuity at $0$ that makes it non-differentiable \cite{schmidt2007smoothl1}.
Several differentiable metrics have been proposed to deal with this issue \cite{cleach_fast_2019, huber_robust_2004}.
In this paper, we study two such costs when applied to \gls{ddp}: the SmoothL1 \cite{fu_retinamask_2019} and Huber differentiable approximations \cite{huber_robust_2004} to $L_1$.
Using such sparsity-inducing cost terms with \gls{ddp} raises two challenges requiring careful consideration: i) ensuring numerical stability/conditioning as efficient implementations assume positive-definiteness of the control Hessian, and ii) trading off achievements of desired tasks with sparsity of solutions through control regularization. Note that these challenges do not arise when using direct transcription/collocation where tasks are enforced with hard constraints and solved using off-the-shelf \gls{nlp} solvers.

\subsection{Contributions}
We study approaches to induce sparsity in optimal control solutions and make the following contributions:
\begin{compactenum}
    \item Introduce sparsity-inducing regularization terms in \gls{ddp}-type solvers.
    \item Compare different strategies for sparsity-inducing regularization, namely SmoothL1, Huber, and Pseudo-Huber loss, in terms of their convergence and numerical stability.
    \item Demonstrate our approach for the swing-up of a cart-pole and for satellite thruster control. Additionally, we demonstrate hardware manipulation experiments using the Valkyrie humanoid robot.
\end{compactenum}
We provide our implementation and evaluations as open source software for reproduction.\footnote{\url{https://github.com/ipab-slmc/sparse_ddp}} A supplementary video is available at \url{https://youtu.be/YMXRZjFsqhc}.

\section{Control Regularization} \label{sec:sparsity-l1}
We begin by reviewing the use of $L_2$ penalties in the optimal control literature. The $L_2$ loss is defined as:
\begin{equation*}
    L_2(x) \triangleq x^2.
\end{equation*}
The $L_2$ loss is used to regularize solutions by penalizing large positive or negative control inputs in the optimal control setting or features in machine learning.

In contrast, the $L_1$ loss is used to penalize solutions for sparsity, and as such, it is commonly used for feature selection in the machine learning community~\cite{tibshirani_regression_1996}. The $L_1$ loss is defined as the absolute value of its argument:
\begin{equation*}
    L_1(x) \triangleq \abs{x}
\end{equation*}
When $L_1$ is used with gradient-based optimizers as a regularization, it drives its argument to exactly $0$ as opposed to small values. This is explained by the condition that the gradients of the regularization parameter and the task cost must be parallel.

Using the $L_1$ loss directly in gradient-based optimization is difficult due to the discontinuity at $x = 0$ where the gradient is undefined. Smooth approximations to the $L_1$ function can be used in place of the true $L_1$ penalty. In this paper, we consider a smooth $L_1$ function from~\cite{fu_retinamask_2019}, which combines $L_1$ and $L_2$ losses, defined as:
\begin{equation}
    L_\text{smooth-l1}(x) = \begin{cases}
        0.5 x^2/\beta &\mbox{if}\ \abs{x} \leq \beta \\
        \abs{x} - 0.5 \beta &\mbox{otherwise}
    \end{cases}
\end{equation}
where $\beta$ defines where the function switches from an $L_1$ to an $L_2$ cost. For small $x \leq \beta$, \textit{SmoothL1} switches to $L_2$, since $L_2$ has a gradient at $0$.

\begin{figure}[t]
    \centering
    \includegraphics[width=.9\columnwidth,trim={0.3cm 0.4cm 0.3cm 0.3cm},clip]{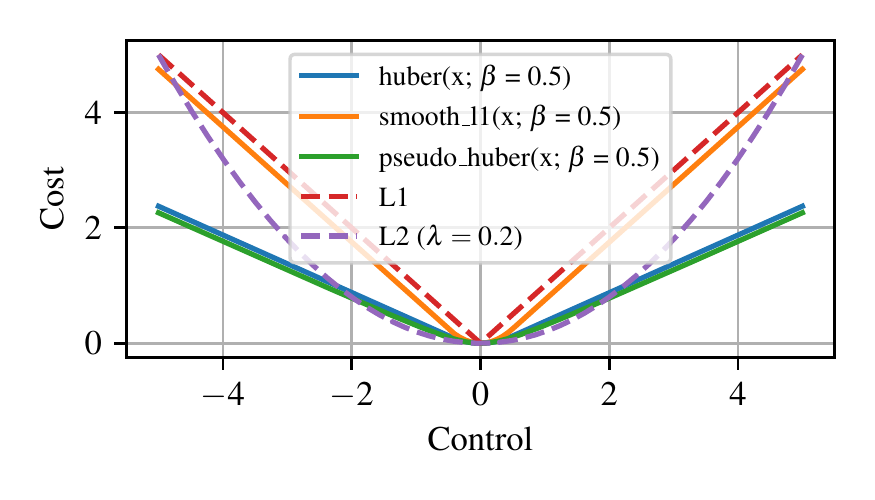}
    \caption{Comparison of the different regularization schemes for the control cost using a range of hyper-parameters: $L_2$, $L_1$, $L_{\text{1-smooth}}$, $L_{\text{Huber}}$, and $L_{\text{pseudo-Huber}}$.}
    \label{fig:compare_different_losses}
\end{figure}

We study another variant of combining $L_1$ and $L_2$ regularization, namely the Huber loss~\cite{huber_robust_2004}:
\begin{equation}
    L_\text{huber}(x) = \begin{cases}
            0.5 x^2 &\mbox{if}\ |x| \leq \beta \\
            \beta(\abs{x} - 0.5 \beta) &\mbox{otherwise}
        \end{cases}
\end{equation}
where $\beta$ is again a shape parameter. The Huber loss has a variable slope, controlled by $\beta$ in addition to mixing $L_1$ and $L_2$. This can be seen in \autoref{fig:compare_different_losses}, where for $\beta = 0.5$ the Huber cost has a lower slope than SmoothL1.

Finally, we consider a smooth approximation to the Huber loss, the Pseudo-Huber loss, as defined in~\cite[Appendix~6]{hartley_multiple_2003}:
\begin{equation}
    L_\text{pseudo-huber}(x) = \beta^2 \bigg(\sqrt{(1 + \frac{x}{\beta}^2)} - 1\bigg).
\end{equation}

We illustrate the considered losses for different settings of their hyper-parameters in \autoref{fig:compare_different_losses}. The shape parameters of the smooth variants control how closely they approximate the true $L_1$ and Huber losses, respectively. The choice of the control regularization and its parametrization has an impact on convergence and sparsity of the output of the optimal control formulation.

\section{Optimal Control}
We consider the robot as a dynamic system described by state $\vx$ composed of generalized coordinates $\vq$ and generalized velocities $\vv$. The system evolves under applied control inputs $\vu$ according to the state transition function $\vx_{t+1} = \vf(\vx_t, \vu_t)$ which incorporates the differential dynamics as well as an integration scheme.
Here, we use a geometric representation of the configuration manifold of floating-base systems ($\mathbb{SE}(3)$) with its geometric integrators along with an energy-conserving symplectic integration scheme of the differential dynamics.

To describe a discrete optimal control problem with a fixed horizon, we additionally specify the integration time step $\Delta t$ and time horizon $T$ and the number of discretization knots $N$. This yields a state trajectory $X = \{ \vx_1, \dots, \vx_N \}$ and control trajectory $U = \{ \vu_1, \dots, \vu_{N - 1} \}$. Tasks and constraints are enforced by minimizing a cost function:%
\begin{equation} \label{eq:cost}
    J(X, U) = h(\vx_N) + \sum_{t = 1}^{N - 1} l(\vx_t, \vu_t).
\end{equation}
Shooting methods in particular minimize $J(\cdot)$ with respect to control inputs only:%
\begin{equation*}
    U^* = \argmin_{U}\ J(X, U)
\end{equation*}
where $U^*$ is the optimal open-loop control trajectory. The corresponding state trajectory is obtained by performing a forward roll-out using the state transition function.

\subsection{Differential Dynamic Programming}
\acrfull{ddp}~\cite{mayne1966ddp,mayne_differential_1973} is a classical method to solve the above unconstrained optimal control problem using Bellman's principle of optimality. \gls{ddp} begins by making a quadratic approximation of the action-value function $Q$ around a reference trajectory $X^{(0)} = \{ \refx_1, \dots, \refx_{N} \}$ and $U^{(0)} = \{ \refu_1, \dots, \refu_{N - 1} \}$. The $Q$-function is defined as:
\begin{equation} \label{eq:q_value}
    Q(\vx, \vu, t) = l(\vx, \vu) + V(\vf(\vx, \vu), t + 1)
\end{equation}
where the value function computes the ``goodness" of state $\vx$, the $Q$-function gives the same quantity for a state and an action. \gls{ddp} minimizes the second-order Taylor expansion of the $Q$-function:
\begin{gather} \label{eq:q-expansion}
\begin{split}
    Q(\vx_t, \vu_t) = Q(\refx_t, \refu_t) + 
    Q_x \delx_t + Q_u \delu_t +\\ \frac{1}{2} \delx_t^T Q_{xx} \delx_t + 
        \frac{1}{2} \delu_t^T Q_{uu} \delu_t + \delu_t^T Q_{ux} \delx_t
\end{split}
\end{gather}
where $\delx_t = \vx_t - \refx_t$ and $\delu_t = \vu_t - \refu_t$ and the subscript notation is shorthand for the partial derivative of $Q$ evaluated at the reference trajectory point for $t\in[1,N]$. In the following, we drop the subscripts to denote way points for readability. We then give the derivatives:
\begin{align*}
    Q_x &= l_x + \vf_x^T V'_x \\
    Q_u &= l_u + \vf_u^T V'_x \\
    Q_{xx} &= l_{xx} + \vf_x^T V'_{xx} \vf_x^T + V'_x \cdot \vf_{xx} \\
    Q_{uu} &= l_{uu} + \vf_u^T V'_{xx} \vf_u^T + V'_x \cdot \vf_{uu} \\
    Q_{ux} &= l_{ux} + \vf_u^T V'_{xx} \vf_x^T + V'_x \cdot \vf_{ux}
\end{align*}
where $V'$ is shorthand for $V(\cdot, t+1)$. The last terms are shorthand for tensor products.

DDP minimizes the quadratic approximation with respect to the new coordinate system. Hence we obtain the local feedback control law:
\begin{equation*}
    \delu_t^* = \argmin_{\delu_t} Q(\cdot) = -Q_{uu}^{-1} (Q_u + Q_{ux} \delx_t)
\end{equation*}
with the feed-forward modification $\vk=Q_{uu}^{-1}Q_u$ and state feed-back term $K=Q_{uu}^{-1}Q_{ux}$.
Since $\delu_t^*$ is the minimum of the $Q$-function, DDP obtains a recursive set of equations for the value function at every time-step:
\begin{gather*}
    V = Q(\vx_t, \vu_t^*) = Q(\refx_t, \refu_t) - Q_u Q_{uu}^{-1} Q_u \\
    V_x = Q_x - Q_{xu} Q_{uu}^{-1} Q_u \\
    V_{xx} = Q_{xx} - Q_{xu} Q_{uu}^{-1} Q_{ux}
\end{gather*}
In order to evaluate these equations, $Q$ and its derivatives are evaluated at the reference trajectory state $\refx_t$ and the optimal control $\delu_t^*$ as calculated above. This is performed in a backward pass from knot $t=N$ to $t=1$. The backward pass is followed by a forward pass in order to obtain the new state sequence $\hat{X} = \{ \hat{\vx}_1, \dots, \hat{\vx}_{N} \}$ and controls $\hat{U} = \{ \hat{\vu}_1, \dots, \hat{\vu}_{N - 1} \}$:
\begin{align*}
    \hat{\vx}_1 &= \refx_1 \\
    \hat{\vu}_t %
        &= \refu_t - \vk_t - K_t (\hat{\vx}_t - \refx_t) \\
    \hat{\vx}_{t + 1} &= \vf(\hat{\vx}_t, \hat{\vu}_t)
\end{align*}

\section[Enforcing Sparsity with L1 and Huber Costs]{Enforcing Sparsity with $L_1$ and Huber Costs}
DDP minimizes a general cost function $J(X, U)$ of the form in \eqref{eq:cost}. In this work, we propose adding an additional cost term for each control input $\vu_t$ that induces sparsity. The new cost function thus becomes:
\begin{equation}
    J(X, U) = h(\vx_N) + \sum_{t = 0}^{N - 1} \big[ l(\vx_t, \vu_t) 
        + \lambda\ l_s(\vu_t) \big]
\end{equation}
where $l_s(\vu_t)$ is one of the sparsity-inducing losses described in \autoref{sec:sparsity-l1} and $\lambda$ is a strength parameter, which controls the relative effects of the regularization loss and the objective loss.

For the cartpole and satellite examples, we use quadratic state costs of the following form:
\begin{align*}
    h(\vx_N) &= (\vx_N - \vx^*)^T Q_f (\vx_N - \vx^*) \\
    l(\vx_t, \vu_t) &= (\vx_t - \vx^*_t)^T Q (\vx_t - \vx^*_t)
\end{align*}
where $Q$ is a diagonal weighting matrix for each of the $NDX$ dimensions of $\vx_t$ and the matrix $Q_f$ is a weighting term for $\vx_N$, respectively.
For the Valkyrie example, we demonstrate the use of nonlinear task cost functions such as end-effector position, stability cost, and joint limits from \cite{exotica}.

The parameters $\lambda$ together with $\beta$ are hyper-parameters and their values define the interaction between the sparsity loss and the optimization criterion (task). We study their effect on the convergence of the solution in detail using the canonical cartpole in the following section.

\section{Effects of Sparsity Loss on Toy Problems} \label{sec:cartpole-pendulum} 
\begin{figure*}[ht]
    \centering
    \begin{subfigure}{0.32\textwidth}
        \includegraphics[width=\textwidth,trim={0.3cm 0.3cm 0.3cm 0.3cm},clip]{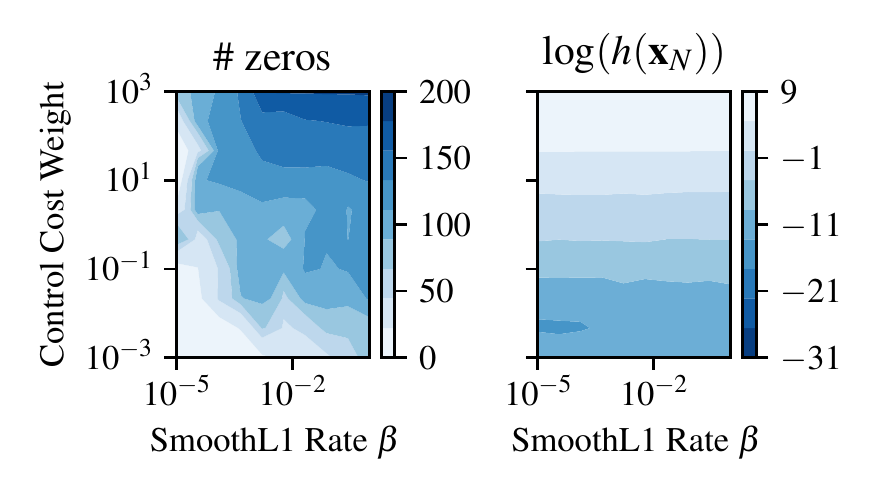}
        \caption{SmoothL1}
        \label{fig:grid_smoothl1}
    \end{subfigure}
    \begin{subfigure}{0.32\textwidth}
        \includegraphics[width=\textwidth,trim={0.3cm 0.3cm 0.3cm 0.3cm},clip]{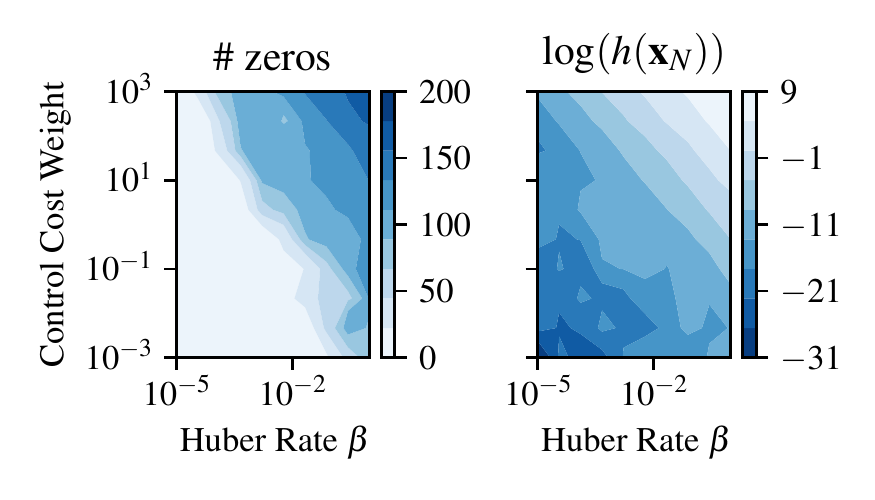}
        \caption{Huber}
        \label{fig:grid_huber}
    \end{subfigure}
    \begin{subfigure}{0.32\textwidth}
        \includegraphics[width=\textwidth,trim={0.3cm 0.3cm 0.3cm 0.3cm},clip]{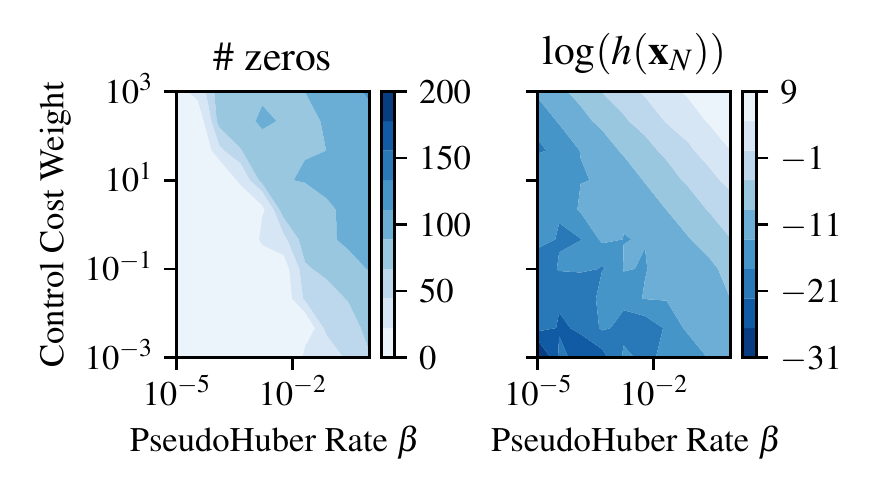}
        \caption{Pseudo Huber}
        \label{fig:grid_pseudo_huber}
    \end{subfigure}
    \caption{Comparison of different sparsity-inducing control cost terms on the cartpole system: Sparsity of solutions and final costs for a grid-search over hyper-parameters $\beta$ (shape) and $\lambda$ (strength). Darker color indicates more sparsity and lower final cost respectively.}
    \label{fig:grid_comparison}
    \vspace{-0.5em}
\end{figure*}

We firstly study the effects of sparsity-inducing costs on a one-dimensional problem---the swing-up of a cartpole, which is a canonical optimal control problem where a pendulum is attached to a cart moving on an infinite friction-less track. The goal is to swing the pendulum upright and move the cart to the origin. The problem is underactuated---the control inputs are linear forces on the cart, whereas the pendulum joint is not controlled. The time horizon is $T=200$ with $\Delta t=0.01\si{\second}$, resulting in a \SI{2}{\second} trajectory. The control limits are $\pm30\si{\newton}$ and $Q_f = 100\ \mathbb{I}_4$, where $\mathbb{I}_4$ is the $4\times4$ identity matrix.

\subsection{Effects of weight and shape parameters on sparsity}
Firstly, we examined the effect of the weighting term $\lambda$ and shape parameter $\beta$ on sparsity. $\beta$ for all functions controls where the switching between an $L_2$ cost (for small $x < \beta$) and an $L_1$ cost (for $x \geq \beta$) occurs. We consider controls to be zero when they are within $[-\beta, \beta]$.

The results of a grid search over $\beta$ and $\lambda$ are in \autoref{fig:grid_smoothl1} for SmoothL1, \autoref{fig:grid_huber} for Huber, and \autoref{fig:grid_pseudo_huber} for PseudoHuber. We plotted both the number of zeros and the final task cost---the latter tells us how close we are to the goal state.

A desired property of sparsity costs is that sparsity should increase with the weight term $\lambda$. As expected, we see a trade-off between sparsity and task cost---the more regularization, the more sparsity, however at a higher task cost. This is in fact what the grid search shows. Another important property we observe is that for lower tolerances $\beta$, a much higher weight is required to achieve sparsity. Thus we can extract a criterion for choosing sparsity---1) pick the largest $\beta$ parameter according to how much noise tolerance the system has and 2) adjust the control cost weight until the desired amount of sparsity is achieved. The noise tolerance of the system is the maximum value of controls beyond which rapid fluctuations can not be tolerated.

\begin{figure}[t]
    \begin{minipage}{.48\textwidth}
        \centering
        \includegraphics[height=1.7in,trim={0.35cm 0.45cm 0.35cm 0.35cm},clip]{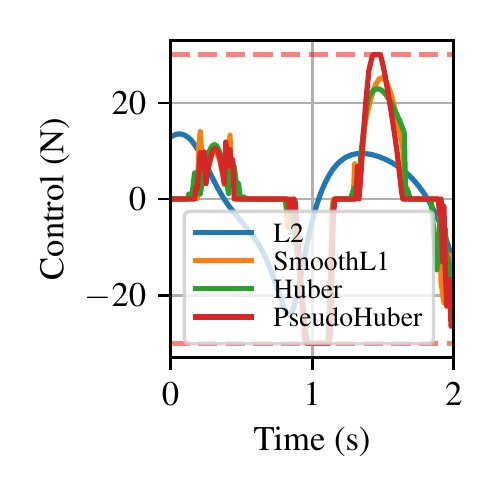}
        \quad
        \includegraphics[height=1.7in,trim={1.55cm 0.45cm 0.35cm 0.35cm},clip]{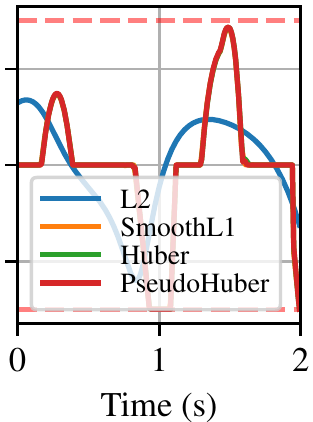}
        \caption{Sparse control solutions for the cartpole system: Left with control artifacts and right with smooth controls.}
        \label{fig:sparse_fail_and_good}
    \end{minipage}
\end{figure}

In the solutions for $\beta=10^{-3}$ in \autoref{fig:sparse_fail_and_good}(left), all solutions achieve a final task cost of less than $10^{-5}$ with $73$, $58$ and $86$ zero controls for SmoothL1, Huber and PseudoHuber, respectively, yet we observe artifacts showing rapid control changes. Increasing the weight (from $7$, $9$, and $0.01$ to $25$, $25$, and $0.023$ for Huber, PseudoHuber and SmoothL1, respectively) can reduce these artifacts, while also increasing sparsity. In \autoref{fig:sparse_fail_and_good}(right) the solutions have $101$, $102$ and $89$ zero controls and produce smoother control profiles. However, this comes with an increase of final state cost from less than $10^{-5}$ to less than $10^{-4}$.

In general, sparsity-inducing costs together with control/actuator limits produce so-called "bang-bang" control. On a real system, aggressive bang-bang control requires the actuator to change the output torque rapidly at each time-step. This is usually not possible due to actuator dynamics, for example, when using electric motors on the cart pole in our example. However, in some domains, namely satellite control, the underlying physical system and actuators are only capable of bang-bang control and this in fact is a desirable property.

\section{Thruster Control for Satellites}
\begin{figure*}
    \centering
    \begin{minipage}{.48\textwidth}
        \centering
        \includegraphics[width=\textwidth,trim={.3cm .45cm .3cm .3cm},clip]{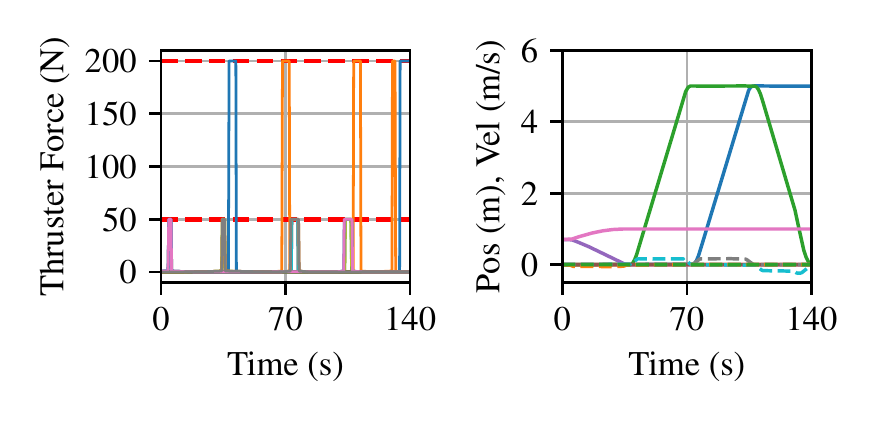}
        \caption{Satellite thruster trajectory (SmoothL1) and corresponding state trajectory.}
        \label{fig:satellite-thruster}
    \end{minipage}
    \hfill
    \begin{minipage}{.48\textwidth}
        \centering
        \includegraphics[width=\textwidth,trim={.3cm .45cm .3cm .3cm},clip]{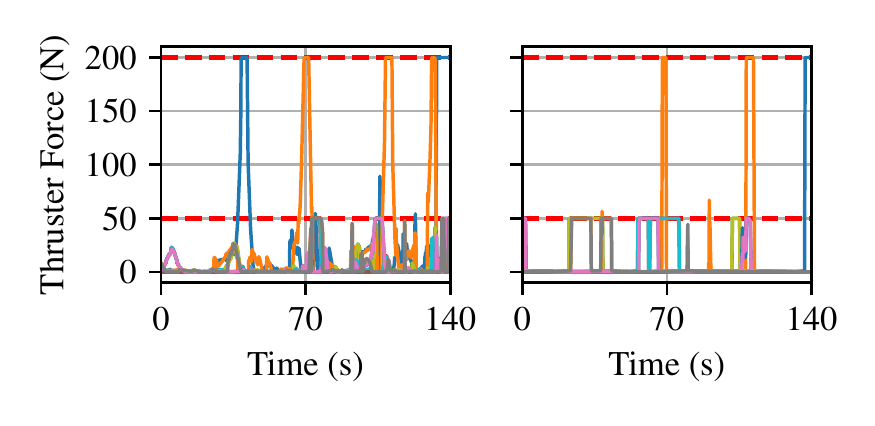}
        \caption{Satellite control trajectories for weights $10^{-5}$ and $10^{-1}$ and a PseudoHuber loss.}
        \label{fig:satellite-weight}
    \end{minipage}
\end{figure*}
We now consider a satellite as a floating rigid body in $\mathbb{SE}(3)$ actuated through forces produced by two primary propulsion thrusters on the front and back of the satellite and four steering thrusters on each of its remaining four sides ($NX=13$, $NU=18$).
We generated a center of mass tracking trajectory with $4$ discrete stages solved in a single optimization problem, as illustrated in \autoref{fig:satellite-maneuver}. The problem was discretized with $\Delta t=0.1$ with $T=1400$ knots for a $140$-second trajectory. Thruster forces were constrained in $[0, 200]~\si{\newton}$ for the front and back and $[0, 50]~\si{\newton}$ for the side thrusters, since there are four for each direction.

The resulting control (thrust) and state trajectories are shown in \autoref{fig:satellite-thruster}. The different colors correspond to different thrusters being activated at the corresponding times. We see thruster peaks at the start and end of each of the stages and reach the corresponding set output forces. The right side of the figure shows the position and linear velocity trajectories in state space.

\subsection{Effects of weight parameters}
We next examine the effects of the different weights on the satellite problem. In \autoref{fig:satellite-weight} we plotted the control trajectories for different weights---$\lambda = 10^{-5}$ and $\lambda = 10^{-1}$. The plot for the correctly tuned $\lambda=10^{-3}$ is in \autoref{fig:satellite-thruster}. We see a similar trend as in the cart-pole system---if the control weight is too small, we observe artifacts in solution space. Tuning the weight produces sparse solutions with no artifacts and the desired bang-bang control profile. Finally, we observe a different result when increasing the weight on the satellite example---this leads to longer thruster bursts on the thrusters with smaller limit, which in fact leads to less sparsity---$23649$ zero controls compared to $24509$ when tuned. This can be explained by the reduction of peaks at $200~\si{\newton}$, which are penalized more, which in turn leads to the solver compensating by switching on the $50~\si{\newton}$ thrusters.

\subsection{Effects on convergence}
\begin{figure}[ht]
    \centering
    \includegraphics[width=.8\linewidth,trim={0.35cm 0.45cm 0.35cm 0.1cm},clip]{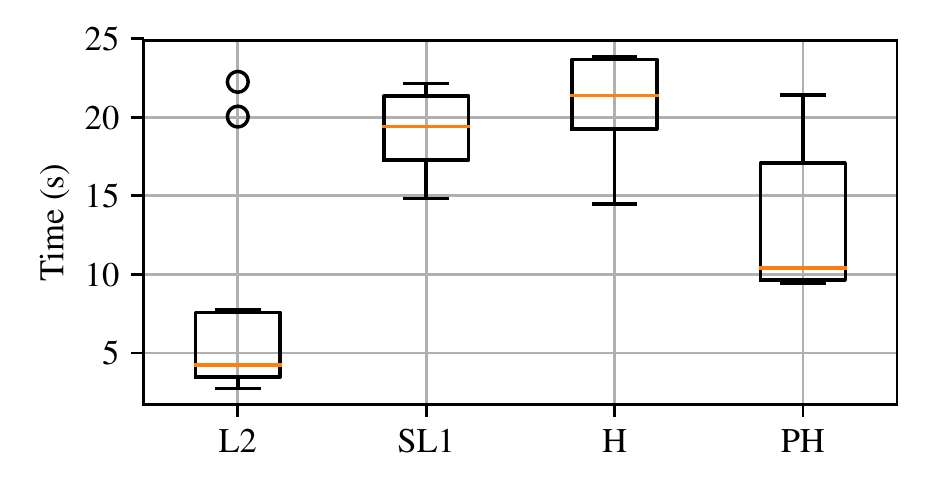}
    \caption{Timing analysis on the satellite example: Using sparse costs leads to slower overall convergence.}
    \label{fig:timing}
\end{figure}

Finally, we examine the effects on convergence for $L_1$ costs. A plot of the time to convergence for all considered cost terms is shown in \autoref{fig:timing}. We computed this over a grid of weights $\lambda \in [10^{-5}, 10^{-1}]$ for $\beta=1$. Generally, time to convergence is increased for all sparse costs with Huber being slowest and Pseudo-Huber fastest to converge on average. This is expected as sparsity-inducing cost terms have less steep gradients further away from zero, where the gradient for an L2 loss would be larger.

\section{Active joint selection for lower-dimensional tasks on redundant systems} \label{sec:valkyrie}

\begin{figure*}[ht]
    \begin{minipage}{0.49\linewidth}
        \includegraphics[width=.9\linewidth,trim={0.3cm 0.45cm 0.3cm 0.3cm},clip]{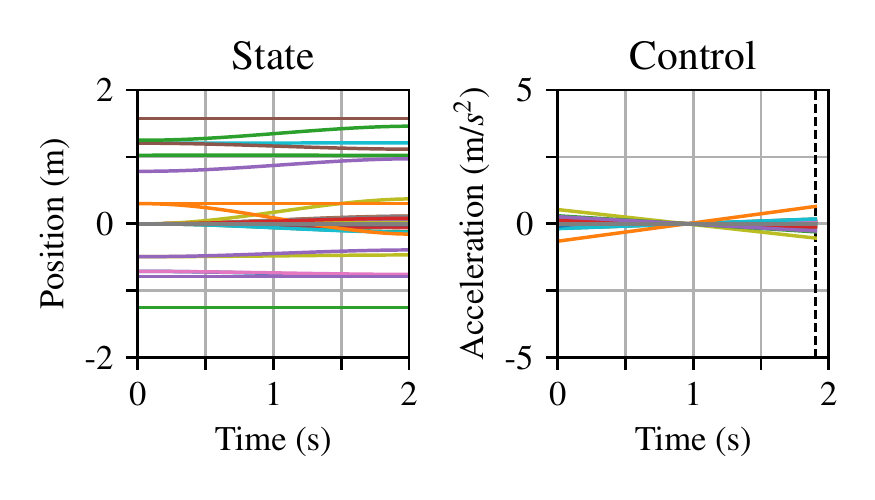}
        \caption{Valkyrie reaching task. $L_2$ uses 25 joints.}
        \label{fig:l2_reach}
    \end{minipage}
    \hfill
    \begin{minipage}{0.49\linewidth}
        \includegraphics[width=.9\linewidth,trim={0.3cm 0.45cm 0.3cm 0.3cm},clip]{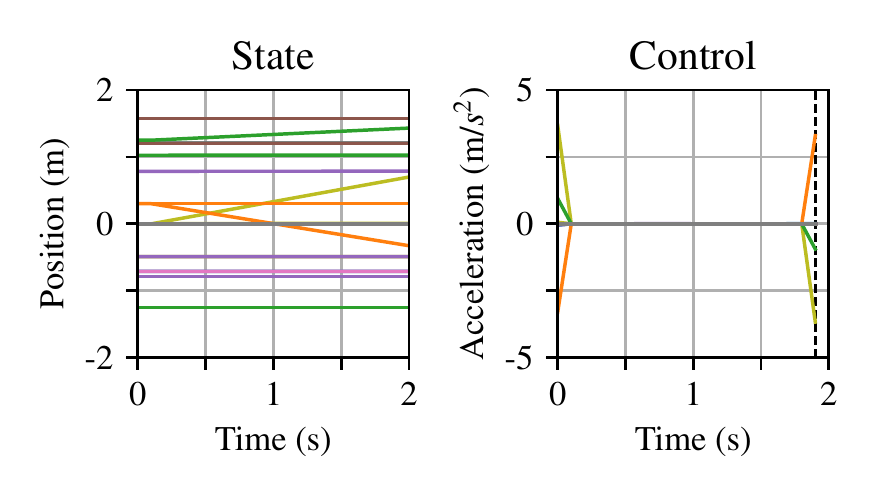}
        \caption{Valkyrie reaching task. Pseudo Huber uses 7 joints.}
        \label{fig:ph_reach}
    \end{minipage}
    \vfill \vspace{1em}
    \begin{minipage}{0.49\linewidth}
        \centering
        \includegraphics[width=0.7\linewidth]{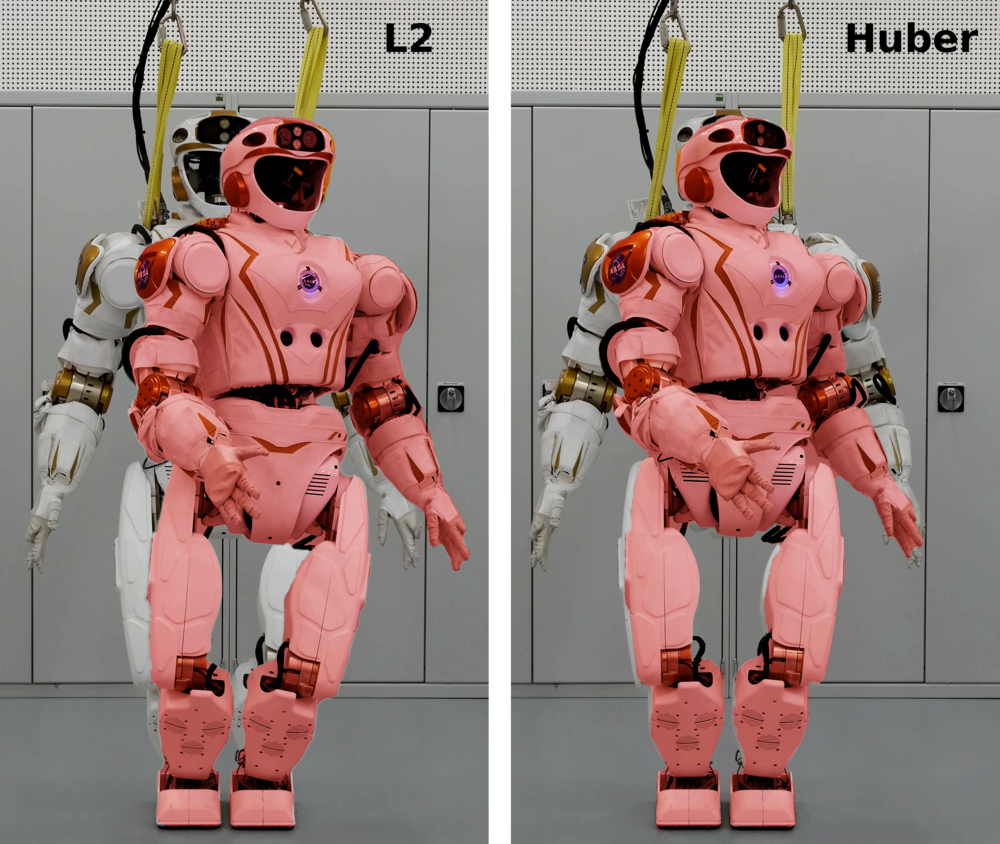}
        \caption{Reaching to a position target: Using $L_2$, the robot reaches the target while moving a majority of the joints by small amounts. A Huber loss induces sparsity improving tracking and maintaining an equal contact force distribution.}
        \label{fig:intro}
    \end{minipage}
    \hfill
    \begin{minipage}{0.49\linewidth}
        \includegraphics[width=0.9\linewidth,trim={5.7cm 0.4cm 0.2cm 0.3cm},clip]{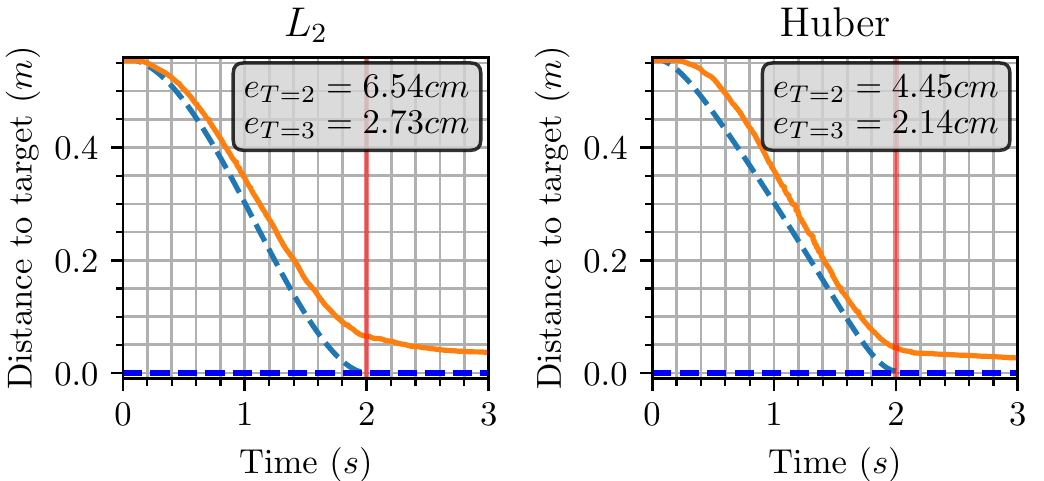} 
        \caption{Tracking results for the Valkyrie reaching task. The red line highlights the end of the commanded trajectory. We highlight the task-space error at the end of the trajectory ($T=2\si{\second}$) and the end of the experiment. This illustrates that the whole-body controller adds a delay to the motion execution, and that the tracking error is lower for the sparsity-induced trajectory.}
        \label{fig:valkyrie_ee_tracking}
    \end{minipage}
    \vspace{-1.2em}
\end{figure*}

We next applied the sparse costs to a reaching task on a 38-\gls{dof} humanoid robot. We use acceleration-based linear system dynamics and nonlinear general task costs. In this case, the three-dimensional reaching task requires only two joints to move (the shoulder and the hip). This is illustrated in \autoref{fig:intro}. The goal is to reach to $\vx^* = [0.5, 0.2, 0.9]$, a point directly in front of the robot at hip-height. We discretized into $T = 20$ knots with $\Delta = 0.1~\si{\second}$ for a \SI{2}{\second} trajectory.

The resulting state and control plots for $L_2$ are in \autoref{fig:l2_reach}. Solving the problem with $L_2$ control regularization produces a solution that moves more joints than necessary. This is easily seen in the corresponding state plots (\autoref{fig:l2_reach} and \ref{fig:ph_reach}), showing the positions and velocities of the joints that move.

Applying a sparsity cost (Pseudo-Huber) to the problem leads to the solver choosing to move only the required joints. The resulting trajectories in \autoref{fig:ph_reach} show this clearly. However, due to the approximation of the $L_1$ costs, numerically the robot is not moving $3$ joints, as is apparent, but rather $7$ have velocities greater than $10^{-3}$. Compared with $L_2$, which moves $26$ joints, this is nonetheless a significant reduction.

Finally, we executed the motion plans on the physical robot.
The trajectories are tracked using an inverse dynamics based whole-body controller. We compared a solution with an $L_2$ cost and a Huber cost. We plot the tracking results (distance of end-effector to target) in \autoref{fig:valkyrie_ee_tracking}. For the Huber sparse trajectory, tracking is better both during and at the end of the trajectory.

\section{Discussion}
We studied the effects of using an $L_1$ cost for the control of dynamic systems using optimal control. Since $L_1$ is not continuously differentiable, we studied three approximations: the SmoothL1, Huber, and PseudoHuber losses. 

On a simple cartpole problem, $L_1$ costs lead to sparsity in control space by making a subset of the controls $0$ and producing peaks that resemble square waves. We analyzed the performance of $L_1$ costs over a grid of values for the shape parameter $\beta$, which thresholds switching between $L_1$ and $L_2$, and the control cost weight $\lambda$. For smaller values of $\beta$ a much larger control weight is required to achieve sparsity. Larger control weights, however, lead to higher task costs. We thus propose picking the largest value of $\beta$ according to the system's noise tolerance and then fitting the weight $\lambda$ until the desired level of sparsity is achieved. We further motivate this approach by showing that relying on sparsity and final task cost alone can lead to non-smooth control trajectories with visible artifacts.

Scaling $L_1$ costs to real-world robots presents new challenges. We successfully applied $L_1$ to a kino-dynamics optimal control problem on the Valkyrie robot to select a subset of active joints for a low-dimensional reaching task. While $L_2$ losses use more joints than necessary, $L_1$ can automatically reduce the number of joints by setting the corresponding controls to $0$. Sparse controls can be in practice tracked better. However, sparse controls also result in bang-bang control with higher commanded accelerations that can damage the hardware.

On the other hand, this is a desired control profile for thruster control for satellites. We were able to achieve thruster-like behavior with $L_1$ costs in order to track a multi-stage center of mass trajectory. We further analyzed the timing performance of $L_1$ costs, showing that there is an increase in convergence time when using $L_1$ costs with Huber being the slowest, followed by SmoothL1 and Pseudo-Huber. Finally, we showed that the weight parameter $\lambda$ has a similar effect for satellite control as it does on the cartpole---low values of $\lambda$ lead to artifacts and high values of $\lambda$ lead to high task costs. In the satellite problem, however, high $\lambda$ did not numerically lead to more sparsity, as it produced longer thruster peaks for the lower control-limit thrusters, instead penalizing the high-limit thrusters more.

Finally, we note that it is important to enforce control limits when using sparsity-inducing losses. For unconstrained methods (DDP \cite{mayne1966ddp}, FDDP \cite{mastalli_crocoddyl_2020}) clamping of applied controls in the forward-pass works in practice but results in slower convergence and often leads to getting stuck in local minima. We tested the losses with active-set control-limited DDP \cite{tassa_control-limited_2014}, and in particular BoxFDDP \cite{mastalli2020boxfddp} in our experiments due to its greater generalization without an initial guess.

Future directions for this research are in time optimization of trajectories using DDP and to address the convergence and control artifacts using regularization strategies.

\clearpage
\bibliographystyle{IEEEtran}
\bibliography{IEEEfull,IEEEconf,optimal_control,sparse}

\end{document}